\documentclass[sigconf]{aamas} 

\doi{BRAG3288}

\usepackage{balance} 
\usepackage{soul}
\usepackage{algorithm}
\usepackage[noend]{algpseudocode}
\usepackage{makecell}

\makeatletter
\gdef\@copyrightpermission{
  \begin{minipage}{0.2\columnwidth}
   \href{https://creativecommons.org/licenses/by/4.0/}{\includegraphics[width=0.90\textwidth]{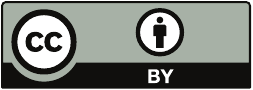}}
  \end{minipage}\hfill
  \begin{minipage}{0.8\columnwidth}
   \href{https://creativecommons.org/licenses/by/4.0/}{This work is licensed under a Creative Commons Attribution International 4.0 License.}
  \end{minipage}
  \vspace{5pt}
}
\makeatother

\setcopyright{ifaamas}
\acmConference[AAMAS '26]{Proc.\@ of the 25th International Conference
on Autonomous Agents and Multiagent Systems (AAMAS 2026)}{May 25 -- 29, 2026}
{Paphos, Cyprus}{C.~Amato, L.~Dennis, V.~Mascardi, J.~Thangarajah (eds.)}
\copyrightyear{2026}
\acmYear{2026}
\acmDOI{}
\acmPrice{}
\acmISBN{}

\acmSubmissionID{<<submission id>>}

\title[AAMAS-2026 Formatting Instructions]{IntentCUA: Learning Intent-level Representations for Skill Abstraction and Multi-Agent Planning in Computer-Use Agents}

\author{Seoyoung Lee}
\authornote{Both authors contributed equally to this research.}
\affiliation{
  \institution{Sookmyung Women's University}
  \city{Seoul}
  \country{Republic of Korea}}
\email{leesy3891@gmail.com}

\author{Seobin Yoon}
\authornotemark[1]
\affiliation{
  \institution{Sookmyung Women's University}
  \city{Seoul}
  \country{Republic of Korea}}
\email{binsong2@sookmyung.ac.kr}

\author{Seongbeen Lee}
\authornote{Corresponding author.}
\affiliation{
  \institution{Sookmyung Women's University}
  \city{Seoul}
  \country{Republic of Korea}}
\email{seongbeen@sookmyung.ac.kr}

\author{Yoojung Chun}
\affiliation{
  \institution{Sookmyung Women's University}
  \city{Seoul}
  \country{Republic of Korea}}
\email{yj.chun@sookmyung.ac.kr}

\author{Dayoung Park}
\affiliation{
  \institution{Sookmyung Women's University}
  \city{Seoul}
  \country{Republic of Korea}}
\email{pdysicist@sookmyung.ac.kr}

\author{Doyeon Kim}
\affiliation{
  \institution{Sookmyung Women's University}
  \city{Seoul}
  \country{Republic of Korea}}
\email{ehdus@sookmyung.ac.kr}

\author{Joo Yong Sim}
\authornotemark[2]
\affiliation{
  \institution{Sookmyung Women's University}
  \city{Seoul}
  \country{Republic of Korea}}
\email{jysim@sookmyung.ac.kr}

\begin{abstract}
Computer-use agents operate over long horizons under noisy perception, multi-window contexts, evolving environment states. Existing approaches, from RL-based planners to trajectory retrieval, often drift from user intent and repeatedly solve routine subproblems, leading to error accumulation and inefficiency. 

We present IntentCUA, a multi-agent computer-use framework designed to stabilize long-horizon execution through intent-aligned plan memory. A Planner, Plan-Optimizer, and Critic coordinate over shared memory that abstracts raw interaction traces into multi-view intent representations and reusable skills. At runtime, intent prototypes retrieve subgroup-aligned skills and inject them into partial plans, reducing redundant re-planning and mitigating error propagation across desktop applications.

In end-to-end evaluations, IntentCUA achieved a 74.83\% task success rate with a Step Efficiency Ratio of 0.91, outperforming RL-based and trajectory-centric baselines. Ablations show that multi-view intent abstraction and shared plan memory jointly improve execution stability, with the cooperative multi-agent loop providing the largest gains on long-horizon tasks. These results highlight that system-level intent abstraction and memory-grounded coordination are key to reliable and efficient desktop automation in large, dynamic environments.
\end{abstract}


\newcommand{\BibTeX}{\rm B\kern-.05em{\sc i\kern-.025em b}\kern-.08em\TeX}

\begin{document}


\pagestyle{fancy}
\fancyhead{}


\maketitle 


\section{Introduction}

\begin{figure*}[t]
  \includegraphics[width=.90\textwidth]{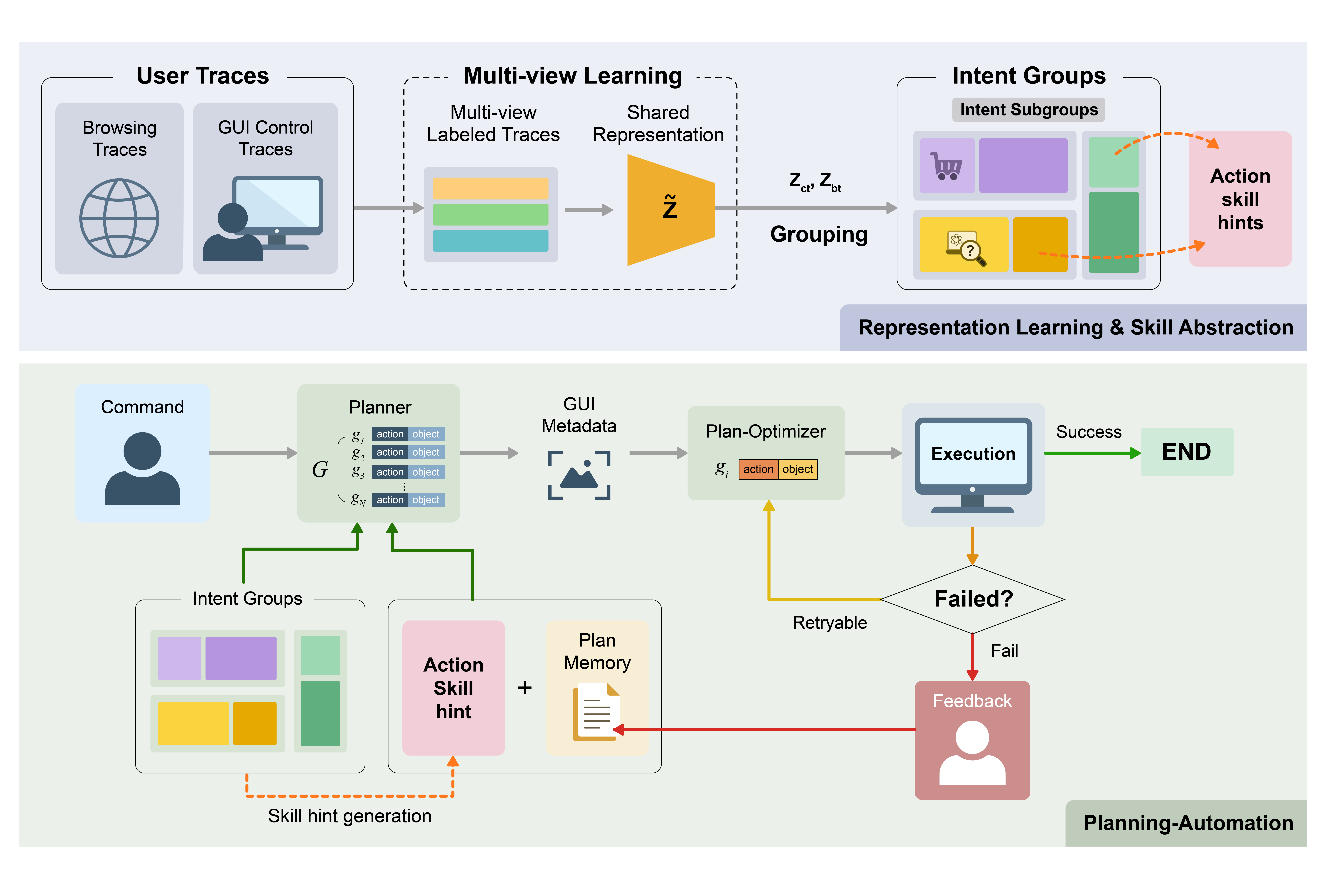}
  \caption{Overview of \textbf{IntentCUA}. \emph{Offline:} raw user traces are multi-view labeled, embedded into a shared intent space, and clustered into intent groups (IG) and subgroups (SG); SG action patterns are converted into parameterized skill schemas (“skill hints”) and stored with their SG in the IG/SG index, while plan memory stores only user-approved global plans (G). \emph{Online:} the Planner/Plan-Optimizer/Critic query and reuse skills; cache-first reuse and template-based gap filling reduce re-planning on long-horizon desktop tasks.}
  \label{fig:main}
\end{figure*}

Rule-based macros and RPA systems enabled early forms of computer-use automation. 
However, they lack adaptability~\cite{tripathi2018learning,krosnick2022parammacros} when compared to recent GUI agents powered by large language models (LLMs) that can interpret screens and generate actions dynamically. 

Research on GUI agents has rapidly expanded, spanning web, mobile, and increasingly desktop environments~\cite{zhang2025largelanguagemodelbrainedgui,sager2025comprehensivesurveyagentscomputer}. 
As highlighted by Tang et al.~\cite{tang2025surveymllmbasedguiagents}, automation across all desktop environments remains particularly challenging due to multi-window operations, OS-level shortcuts and APIs, and the need to adapt to frequent updates and complex software ecosystems. 
Within such environments, achieving robust long-horizon planning and managing multi-context workflows emerge as central challenges that current systems have yet to overcome.

Recent multi-modal agents attempt to address these challenges by perceiving screens and generating actions with large models~\cite{anthropic2024computeruse,yang2023mm}. 
However, robust long-horizon automation across heterogeneous desktop applications remains unresolved~\cite{sager2025ai, tang2025surveymllmbasedguiagents}. 
We identify two recurring failure modes: 
(i) plans spanning multiple substeps often drift from the original intent and redundantly re-solve already completed routines~\cite{redis2024skill,rebmann2024recognizing}, 
(ii) local perception errors accumulate and lead to cascading retries~\cite{zhang2024dynamic,lu2024omniparser,li2024ferret,hong2024cogagent}.
These factors collectively hinder robust long-horizon planning, as agents frequently fall into inefficient and repetitive re-planning cycles.
Actions are often retried or nullified when context drifts, leading to prolonged latency and unstable completion rates.

To address these limitations, we bridge user interaction and multi-agent planning. Rather than simply replaying trajectories or storing textual reflections~\cite{shinn2024reflexion}, we transform interaction traces into labeled units, induce generalized skills from sub-intent clusters, and learn multi-view representations across environment, action, keyword, and description.

These skills are organized hierarchically in a plan memory and retrieved via semantic search during planning, which supports cross-application transfer and helps stabilize long roll-outs. At runtime, intent prototypes are projected into a shared embedding space, where centroid-based retrieval augments partial plans with relevant skills.

In end-to-end evaluations, IntentCUA achieves a 74.83\% task success rate with a Step Efficiency Ratio (SER) of 0.91, outperforming both RL-based (UI-TARS-1.5~\cite{ui-tars-15-seed}) and trajectory-centric (UFO$^2$~\cite{zhang2025ufo2}) baselines in success rate, efficiency, and latency. 
Ablation studies confirm that multi-view intent abstraction and shared plan memory jointly improve execution stability, with the cooperative multi-agent loop providing the largest gains on long-horizon tasks. These results indicate that system-level intent abstraction and memory-grounded coordination are central to reliable desktop automation.

Our contributions are summarized as follows:
\begin{enumerate}
\item We propose \textbf{IntentCUA}, a multi-agent computer-use framework that stabilizes long-horizon execution through intent-aligned plan memory and coordinated planning.

\item We introduce a trace-to-skill abstraction pipeline that learns multi-view intent representations and induces hierarchical, reusable skills from raw user interaction traces.

\item We design a planning-time memory mechanism that retrieves subgroup-aligned skills to augment partial plans, reducing intent drift and redundant re-planning in dynamic desktop environments.

\item We demonstrate through extensive ablations and end-to-end evaluations that intent abstraction and memory-grounded coordination significantly improve execution stability, efficiency (SER 0.91), and task success (74.83\%) on complex desktop workflows.
\end{enumerate}

In summary, IntentCUA shows that intent-level abstraction and memory-grounded multi-agent coordination are key to stabilizing long-horizon desktop automation in large, dynamic environments.
This automation is made possible by a robust planning policy that maintains coherence and efficiency across extended sequences.


\section{Related Work}
\subsection{Desktop and GUI Automation Agents}

GUI automation spans web, mobile, and desktop domains. Web agents such as WebArena and WebVoyager operate under structured DOM feedback~\cite{zhou2023webarena,he2024webvoyager}, but real desktop environments lack such schema-level constraints and require cross-application coordination.

Desktop benchmarks like OSWorld~\cite{xie2024osworldbenchmarkingmultimodalagents}  tasks are typically long-horizon, requiring stable execution 10–20 sequential steps. This makes execution latency-sensitive and error-prone. As step count increases, local perception errors compound and agents often enter loops of repeated or failed actions.

Recent desktop agents such as UI-TARS~\cite{qin2025ui}, UFO~\cite{zhang2024ufo}, ScreenAgent~\cite{niu2024screenagent} extend vision-language models with planner–critic loops. However, surveys highlight a persistent challenge: determining actions that align with specific user contexts and preferences in dynamic, interruption-prone interfaces~\cite{tang2025surveymllmbasedguiagents}. Even with improved GUI grounding~\cite{lu2024omniparser,hong2024cogagent,jiang2025iluvui}, intent drift and redundant re-planning remain common in long-horizon workflows.

These findings indicate that stable long-horizon planning, rather than perception alone, remains the key bottleneck for reliable desktop automation.

\subsection{Agents Leveraging Interaction Traces}
One approach to addressing long-horizon instability is to learn directly from large-scale interaction traces.

Macro-mining and process-mining techniques cluster demonstrations into recurrent procedures or labeled schemas~\cite{huang2024automatic,fani2023llms,choi2022enabling}. Large-scale corpora such as OS-ATLAS support perception pretraining across millions of GUI elements~\cite{wu2025osatlas}. Offline reinforcement learning has also been explored for device agents~\cite{song2023navigating}, while systems such as AppAgentv2~\cite{li2025appagentv2advancedagent}, AgentBank~\cite{song2024agentbank}, and UI-TARS-1.5~\cite{ui-tars-15-seed} leverage hierarchical feedback or large-scale trajectory tuning to improve control robustness.

These works demonstrate that interaction traces improve policy generalization and low-level stability. However, most approaches operate at the trajectory or action level, emphasizing replay or large-scale tuning rather than structured intent abstraction. As a result, redundancy and error accumulation often persist in long-horizon execution~\cite{redis2024skill}, and reliance on controlled environments or explicit reward signals limits applicability to open-ended desktop workflows~\cite{sager2025comprehensivesurveyagentscomputer}.

\subsection{Plan Memory, Intent Identification, and Skill Abstraction}
A complementary direction enhances robustness through memory retrieval and skill abstraction.

Memory-based methods such as Reflexion~\cite{shinn2024reflexion}, Conversational Memory~\cite{wang2023conversational}, and Contextual Experience Replay~\cite{cer2025} retrieve prior trajectories, manuals, or reflections to guide future decisions~\cite{cai2023low}. Skill-level prompting approaches such as SkillAct~\cite{liu2024skillact} show that abstracted routines can improve interactive performance, while UFO$^2$~\cite{zhang2025ufo2} manages app-specific demonstrations as reusable references.

Parallel work investigates intent recognition from UI logs~\cite{rebmann2024recognizing,li2020mapping} and representation learning of screens (e.g., Screen2Vec~\cite{li2021screen2vec}, Aria-UI~\cite{yang2024aria}), while GUI grounding methods reduce perceptual ambiguity~\cite{lu2024omniparser,li2024ferret}. More recent systems explore adaptive planning and dependency modeling from demonstrations~\cite{zhang2024dynamicplanningGUI,yin2025cognitivedependencies}.

Despite these advances, structured and hierarchical skill abstractions that remain transferable across heterogeneous desktop workflows are still relatively underexplored. As a result, maintaining stable long-horizon execution under dynamic user contexts continues to be an active area of research.
Our approach complements these directions by learning multi-view intent representations that integrate environment, action, and description signals. Skills are stored as hierarchical intent prototypes in plan memory and retrieved to augment partial plans, supporting stable long-horizon execution ~\cite{song2024visiontasker,gao2023assistgui}.


\section{Intent-level Representation Learning \& Skill Abstraction}
\label{sec:3}

\subsection{Intent-level Representation Learning}
\label{sec:3.1}

In this section, we describe how raw user traces are transformed into unified intent-level representations that can be clustered and later abstracted into reusable skills. 

\begin{figure}[t]
    \centering
    \includegraphics[width=\linewidth]{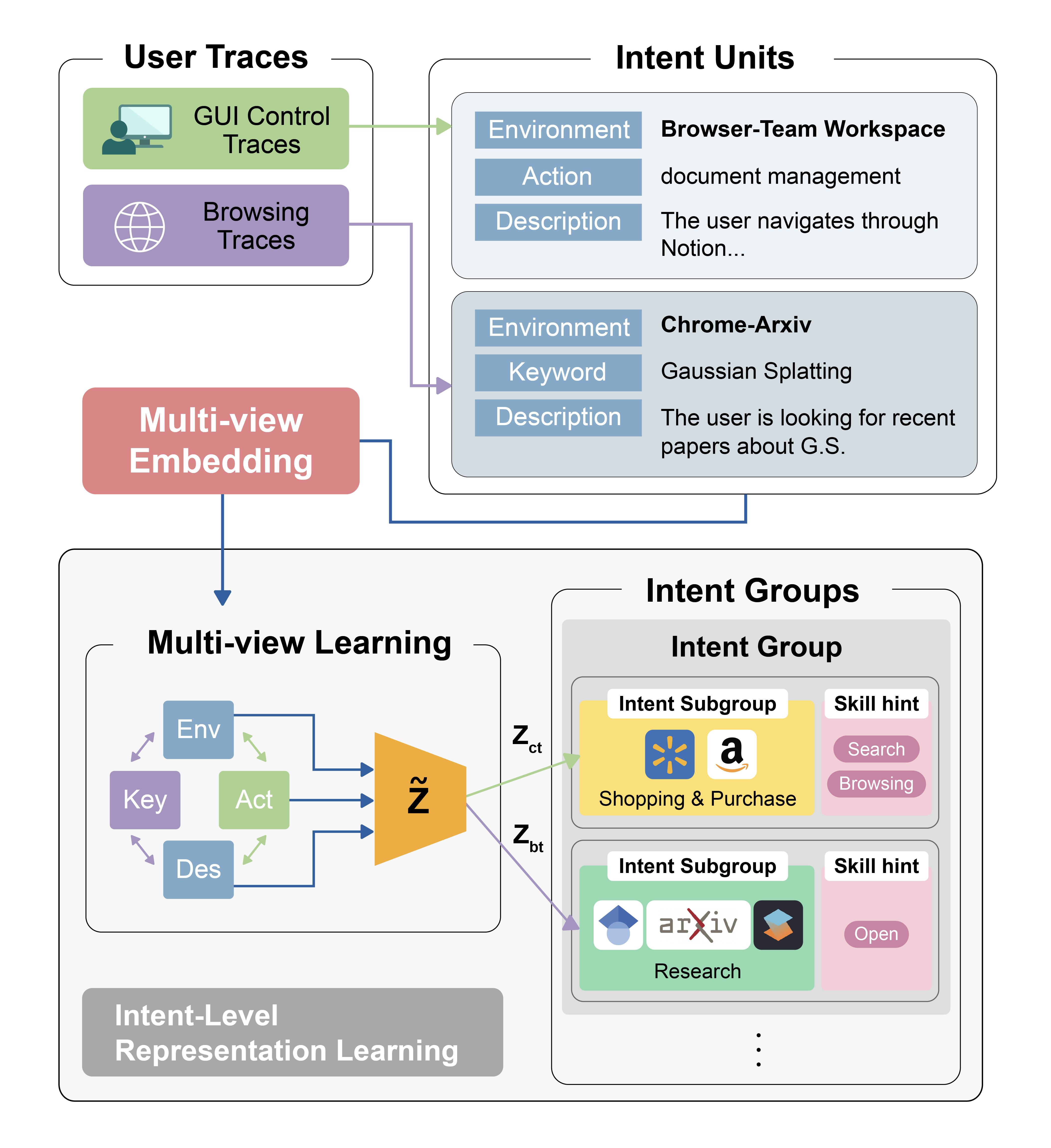}
    \caption{Multi-view intent representation. Control traces use [E,A,D], browsing traces [E,K,D]. A multi-view encoder aligns views into a shared space, inducing environment-centric IG and finer SG. SG centroids enable retrieval, and SG action patterns are converted into parameterized skill schemas (“skill hints”) with verb–argument structure for planning.}
    \label{fig:multiview_clustering}
\end{figure}

As shown in Figure~\ref{fig:multiview_clustering}, each user trace is first labeled across four views: environment ($E$ with instances $e_i$), action ($A$ with instances $a_i$), keyword ($K$ with instances $k_i$), and description ($D$ with instances $d_i$), 
where $i$ indexes the sequential intent units that together compose a user’s interaction trace.  
Each view $v \in \{E,A,K,D\}$ is represented as an embedded textual vector, capturing its semantic content.  
Control traces produce intent units, $u_i$, of the form $[e_i,a_i,d_i]$, while browsing traces yield $[e_i,k_i,d_i]$. 
Formally, let $x_i^{(v)}$ denote the feature representation of intent unit $u_i$ in view $v \in \{E,A,K,D\}$. 
A multi-view encoder $\phi(x^{(v)})$ maps these view-specific features into a single shared representation $z_i$:
\begin{equation}
z_i = \phi\!\left((x^{(v)}_{i})_{v \in V}\right)\in \mathbb{R}^d,
\quad V \subseteq \{E,A,K,D\}.
\label{eq:zu-def}
\end{equation}

Building on prior multi-view clustering objectives ~\cite{9577930}, we train the model to ensure that representations from heterogeneous views are (i) contradistinctively aligned, (ii) cross-view predictive, and (iii) reconstructible.  
The overall loss is defined as the weighted sum of these three components, as shown in Equation ~\ref{eq:mv_loss_main}:
\begin{equation}
\mathcal{L}
= \mathcal{L}_{\mathrm{con}}
+ \lambda_{\mathrm{pred}}\,\mathcal{L}_{\mathrm{pred}}
+ \lambda_{\mathrm{rec}}\,\mathcal{L}_{\mathrm{rec}}
\label{eq:mv_loss_main}
\end{equation}

where $\mathcal{L}_{\mathrm{con}}, \mathcal{L}_{\mathrm{pred}}, \mathcal{L}_{\mathrm{rec}}$ are cross-view contrastive loss, dual prediction loss, and within-view reconstruction loss, respectively. 

$\mathcal{L}_{\mathrm{con}}$, enforces consistency between embeddings from different views of the same intent unit while separating embeddings from different instances:
{\small
\begin{equation}
\mathcal{L}_{\mathrm{con}}
=
\frac{1}{|P(V)|}\!\sum_{(p,q)\in P(V)}
\Bigg[
-\frac{1}{N}\sum_{i=1}^{N}
\log
\frac{\exp\!\left(\langle z_{i}^{(p)},\, z_{i}^{(q)}\rangle/\tau\right)}
{\sum_{j\neq i}\exp\!\left(\langle z_{i}^{(p)},\, z_{j}^{(q)}\rangle/\tau\right)}
\Bigg]
\label{eq:mv_loss_con}
\end{equation}
}

where, $P(V)$ denotes the set of all ordered positive view pairs $(p,q)$ within the selected view set $V$.  
The temperature parameter $\tau$ controls the sharpness of the contrastive distribution, whereas $N$ denotes the number of intent units sampled in a minibatch.

$\mathcal{L}_{\mathrm{pred}}$ introduces two projection heads $G_{p\!\to\! q}$ and $G_{q\!\to\! p}$ that learn to predict the embedding of one view from another.  
Their averaged mapping $G=(G_{p\!\to\! q}+G_{q\!\to\! p})/2$ acts as a symmetric predictor encouraging cross-view consistency—ensuring that one view can reconstruct another within the latent space.  
The scalar coefficient $\lambda_{\mathrm{pred}}$ balances this term with the others in Equation ~\ref{eq:mv_loss_main}.

\begin{align}
\mathcal{L}_{\mathrm{pred}}
&=
\frac{1}{|P(V)|}\!\sum_{(p,q)\in P(V)}\!
\frac{1}{2N}\sum_{i=1}^{N}
\Big[
\|G_{p\!\to\! q}(z_{i}^{(p)})-z_{i}^{(q)}\|_2^2 \nonumber\\
&\hspace{1.4cm}
+ \|G_{q\!\to\! p}(z_{i}^{(p)})-z_{i}^{(q)}\|_2^2
\Big]
\label{eq:mv_loss_pred}
\end{align}

Finally, $\mathcal{L}_{\mathrm{rec}}$ ensures that the shared embedding ${z_i}$ retains view-specific semantics by reconstructing each original feature $x_v^{(u)}$ through a decoder $g_v(\cdot)$.  
The weight $\lambda_{\mathrm{rec}}$ determines the relative strength of this reconstruction constraint within the total loss in Equation ~\ref{eq:mv_loss_main}.
\begin{equation}
\mathcal{L}_{\mathrm{rec}}
=
\frac{1}{|V|\,N}\sum_{v\in V}\sum_{i=1}^{N}
\|g_v(z_{i}^{(v)})-x_{i}^{(v)}\|_2^2
\label{eq:mv_loss_rec}
\end{equation}

Together, these three objectives jointly align, predict, and reconstruct heterogeneous views, yielding a coherent embedding space where intent-level semantics are preserved.  
The resulting representation $z_i$ compactly captures user intent across multiple modalities. 
We then organize $\{u_i\}$ hierarchically by $z_i$: first into higher-level intent groups ($IG$) driven by environment context, and then into subgroups ($SG$) based on action/keyword and description. 
These $SG$ representations provide the foundation for extracting recurrent action patterns and constructing abstract skills, as described in Section~\ref{sec:3.2}.
Details of the encoder are in Appendix ~\ref{app:encoder}.

\subsection{Skill abstraction based on Intent Subgroups}
\label{sec:3.2}
Given the per-unit embeddings $z^{(u)}$ from Section~\ref{sec:3.1}, we organize intents into a two-level index for planning. First, we run HDBSCAN~\cite{campello2015hierarchical} over $\{z^{(u)}\}$ to obtain higher-level \emph{intent groups} ($IG$) driven primarily by environment/context. Within each $IG$, a second HDBSCAN partitions units into finer \emph{subgroups} ($SG$) using action/keyword and description signals.
For every $SG$, we compute and store its centroid $c_{SG}$ in the same embedding space. At retrieval time, we rank subgroups by the cosine similarity between $c_{SG}$ and a query \emph{intent prototype}. \textbf{Retrieval index:} for each $SG$ we store (i) the centroid $c_{SG}$ (for cosine-based ranking); (ii) top-$k$ \emph{representative traces} preselected by proximity to $c_{SG}$ and reranked at query time by similarity to the intent prototype; and (iii) a \emph{support} count for $SG$ (defined below) used to prefer stable patterns during planning.
A detailed ablation on the representation loss and $IG/SG$ Gating are reported in Appendix ~\ref{ablation}.

We then consolidate the low-level action traces inside each $SG$ into a reusable skill. Each intent unit $u_i$ in a subgroup is linked to a low-level action sequence: \( M_i = (a_1, a_2, \dots, a_m)_i . \). To make traces comparable, we map every atomic action $a_t$ to a pair \emph{[verb predicate, typed argument fields]} by applying an alias map $\Phi$ that collapses surface variants to a fixed predicate and a fixed set of typed fields. For example, "focus URL bar" and "open web site" $\rightarrow$
{verb=press, arg=address\_bar}, {verb=text\_input, arg=address\_bar, text:"https://example.com"}.
Here, a \emph{verb signature} is the ordered list of canonical predicates in a trace, and an \emph{typed argument field} is a placeholder (e.g., \texttt{<url>}, \texttt{<query>}, \texttt{<file\_path>}) that will be bound at runtime.

We collect candidates \( \mathrm{sg}_{\mathrm{skills},k} = \{\, M_i \mid u_i \in SG_k \,\} \) and induce a \emph{skill prototype} as the medoid under a signature-level dissimilarity \(d{\mathrm{sig}}\) over verb-predicate sequences.  The function $d_{\mathrm{sig}}$ is computed on the canonicalized verb-predicate sequences (after applying $\Phi$), comparing action patterns at the predicate level while deferring literal-argument handling to the parameterization stage.
Let $\mathcal{A}$ denote the verb-predicate alphabet. 
Thus the subgroup’s \emph{skill prototype} is defined as:
\begin{equation} 
\mathcal{S}_{SG} 
= \arg\min_{a \in A^{*}} 
\;\sum_{s \in {sg}_{skills}} \mathrm{d}_{\mathrm{sig}}(a, s) 
\label{eq:sg-medoid-sig} 
\end{equation}

Next we convert $\mathcal{S}_{SG}$ into a reusable, \emph{parameterized schema, skill hint}: a verb-predicate sequence together with a typed argument structure. This conversion (i) replaces literal values with typed parameters (the runtime-filled fields), (ii) removes incidental or recovery-specific steps that do not affect goal attainment, and (iii) enforces canonical predicate and field names via $\Phi$. We refer to this parameterized schema as a \emph{skill hint}. Both the skill hints $\{\mathcal{S}{SG}\}$ and the representative traces are stored in plan memory and retrieved at planning time.

If multiple predicate sequences are well supported, we keep several schemas ranked by their \emph{support}, where support counts subgroup members whose similarity to $\mathcal{S}_{SG}$ exceeds a fixed threshold $\tau$. \emph{Representative traces} are the top-$k$ members minimizing $d{\mathrm{sig}}$ to $\mathcal{S}{SG}$ and serve as concrete exemplars. At planning time, when a retrieved plan is only a partial match, we perform \textbf{gap filling}: we instantiate the selected \emph{skill hint} $\mathcal{S}{SG}$ with current-context bindings and insert the resulting steps to complete the missing plan units/steps (Section~\ref{sec:4.1}).

\section{Intent-aware Planning \& Feedback Memory}

Building on the intent-level DB introduced in Section~\ref{sec:3}, we now focus on how these abstractions are leveraged during planning and execution. This section details the end-to-end workflow in which the Planner, Plan-Optimizer, and Critic cooperate through plan memory to compose, refine, and verify long-horizon automation. The Planning-Automation part in figure~\ref{fig:main} represents the overall automation process after the user request.

\subsection{Planning with Plan Memory and Skill Hints}
\label{sec:4.1}

\begin{figure}[b]
    \centering
    \includegraphics[width=\linewidth]{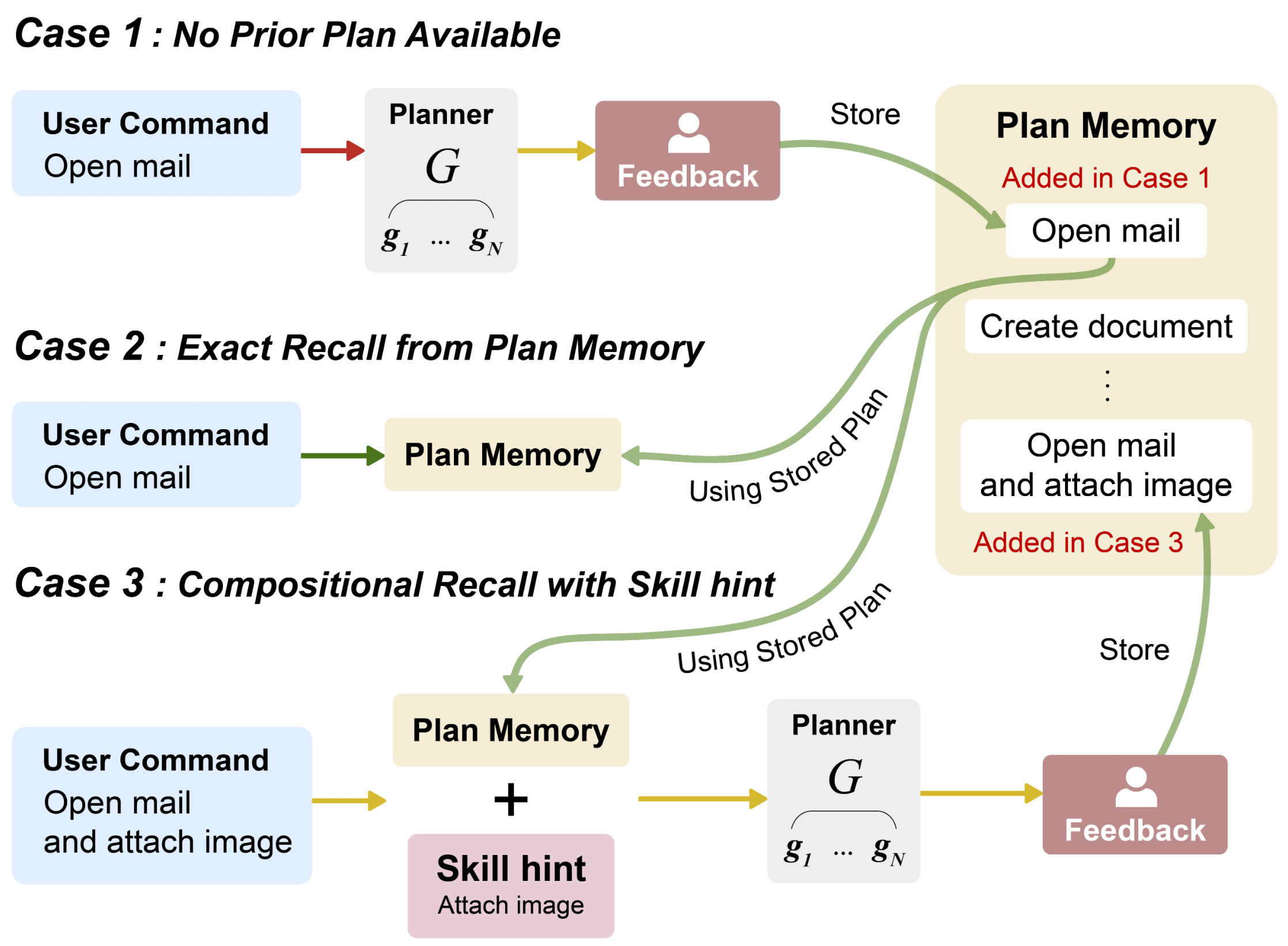}
    \caption{Cache-first planning with plan memory. A query intent is gated by IG and ranked over SG. Case 1 (miss): synthesize a plan from retrieved skill templates. Case 2 (hit): reuse the stored plan. Case 3 (partial): align to the nearest plan and fill gaps with SG-derived skill hints, reducing retries.}
    \label{fig:plan_memory}
\end{figure}

This section explains how the Planner agent composes a high-level plan $G$ for a given command;
Figure~\ref{fig:plan_memory} summarizes three pathways: cache miss synthesis (Case 1), direct reuse on exact hit (Case 2), and reuse-with-injection on partial hit (Case 3).
At a high level, Planner consults $IG/SG$ centroids in the shared embedding space, plan-memory entries, and $M(u)$ sequences; missing spans are completed with normalized $S_{SG}$ templates.

When no suitable plan exists in plan memory (Figure~\ref{fig:plan_memory}, Case~1), the Planner embeds the command into the shared space, gates by the active $IG$, and ranks candidate $SG$ by centroid similarity. Plan memory stores previously synthesized global plans \(G\) that received \emph{binary} user approval. Plans not approved are discarded. For each plan-unit slot in the intent prototype, it selects the nearest $SG$, retrieves the top-2 intent units ($u_i$) from that $SG$, and uses the GPT-4o model~\cite{openai2024hello} to generate the slot’s steps conditioned on the retrieved $M_i$ sequences. A \emph{plan unit} is a contiguous block of low-level steps in the global plan \(G\) that achieves one subgoal. A \emph{plan-unit slot} is the placeholder for such a block in the intent prototype that the Planner must populate with executable steps.

Concatenating all slots yields a high-level plan $G=\{g_1,\dots,g_n\}$, where each $g_j$ expands to a contiguous, execution-ordered list of low-level actions. After the user reviews the generated plan and provides optional feedback, we incorporate the edits and then store $G$ in plan memory for future reuse. Concretely, $G$ is materialized as \emph{plan units}—intent-prototype–level chunks of the $G$ derived from the initial user command:
\noindent\texttt{Plan Unit 1: [step\_1, step\_2, ...]} 
\quad
\texttt{Plan Unit 2: [step\_$k$, ..., step\_$\ell$]} 
\quad
\texttt{...}

When a high-similarity plan is found in active $IG$ (Figure~\ref{fig:plan_memory}, Case 2),
the stored $G$ is retrieved and its slots are bound to the current context;
because no gaps remain, the planner skips the synthesis and executes the plan as-is.

When only a partial match is found (Figure~\ref{fig:plan_memory}, Case 3), the closest stored plan $G$ is aligned with the intent prototype. Insert missing plan units or steps by injecting the matched subgroup’s $S_{SG}$ with current-context bindings (gap filling), after which the finalized \(G\) is executed.
This cache-first pipeline reduces re-planning and stabilizes long-horizon execution by combining centroid gating, plan reuse, and hint-based gap filling.


\subsection{Optimizing steps by memory \& feedback loop}
\label{sec:4.2}

Given a finalized plan $G$, execution shifts into a cooperative loop between the Plan-Optimizer and the Critic Agent, utilizing the Plan Memory.
The Plan-Optimizer refines each plan unit by referencing similar traces stored in the memory, dynamically adapting its substeps to current screen contexts.
The Critic, in turn, monitors the execution and provides immediate feedback signals—\textsf{success}, \textsf{retryable}, or \textsf{blocked}—to correct local deviations or trigger partial replanning when necessary.

For each plan unit $pu$, we compute its representation $z^{(pu)} \in \mathbb{R}^d$ (Section~\ref{sec:3.1}) and compare it with subgroup centroids $\{c_{SG}\}$ from plan memory (Section~\ref{sec:3.2}). 
If a subgroup is relevant, its traces are injected as \emph{hints} into the Plan-Optimizer to guide step execution. 
After each unit, the Critic evaluates the post-execution state and returns $q\in\{\textsf{success},\textsf{retryable},\textsf{blocked}\}$ with an observation $o$. 
If q is \textsf{retryable}, the Plan-Optimizer is re-invoked on the latest state $s^{\text{after}}$ with observation observation of the current GUI context ($o$) to produce an adjusted subplan $g'_{\text{new}}$, which updates $G$ before re-execution. 

\begin{algorithm}[t]
\caption{Execution of the Plan utilizing Memory and Feedback Loop}
\label{alg:execution-memory-feedback}
\begin{algorithmic}[1]
\Require  Final global plan $G$ with plan units $PU=\{pu_1,\dots,pu_M\}$, where each $pu$ is an ordered list of steps;
        subgroup collection $SG$ with centroids $c_{SG}\in\mathbb{R}^d$ (representation space from Section~\ref{sec:3.1});
        for every $pu\in PU$, its representation $z^{(pu)}\in\mathbb{R}^d$ (precomputed via the encoder in Section~\ref{sec:3.1});
        action space $A$.
\Ensure Execution outcome

\For{\textbf{each} plan unit $pu$ in $PU$}
    \State \textit{hint} $\gets$ \Call{InjectHint}{pu, $SG$, $z^{(pu)}$}
    \For{\textbf{each} step $g$ in $pu$}
        \State $s \gets$ GUI Grounding of the current screen
        \State $(a, g', o) \gets$ \Call{Plan-Optimizer}{$s, g, G, pu, o, \textit{hint}$}
        \State Execute action $a$ in the current GUI context
    \EndFor

    \State $s^{\text{after}} \gets$ GUI Grounding of the screen after finishing $pu$
    \State $(q, o) \gets$ \Call{Critic}{$pu, G, s^{\text{after}}$}
    \If{$q == \textsf{success}$}
        \State \textbf{continue} to next $pu$
    \ElsIf{$q == \textsf{retryable}$}
        \State $(a, g'_{\text{new}}, o) \gets$ \Call{Plan-Optimizer}{$s^{\text{after}}, g, G, pu, o, \textit{hint}$}
        \State Apply $g'_{\text{new}}$ to adjust the prior $g'$
    \Else
        \State \Return $(G, \textsf{BLOCKED})$
    \EndIf
\EndFor

\State \Return $(G, \textsf{SUCCESS})$
\end{algorithmic}
\end{algorithm}

As shown in Algorithm \ref{alg:execution-memory-feedback}, the Planner hands over the plan units to the Plan-Optimizer, which integrates hints from prior traces to refine step execution.
The Critic then decides whether to proceed, request an adjustment, or terminate.
Through this memory-guided collaboration, specialized agents coordinate to minimize redundant re-planning and improve robustness by reusing traces that previously led to success.

GUI Grounding refers to the process of enumerating all actionable GUI components on the current screen, similar to the screen parsing method used in UFO \cite{zhang2024ufo}. The resulting state $s$ includes such component data together with summary metadata, composed of window title, panel names and component counts captured from the environment.
Each step $g$ denotes an individual operation in the global plan $G$, composed of an action $a$ (e.g., click, text input, open) and its corresponding object targets; thus $a$ specifies the interaction primitive, whereas $g$ represents the full executable tuple $(a,\text{object})$.

The \textsc{InjectHint} function searches the plan memory for previous plan units whose representations $z^{(pu)}$ are most similar to the current one, and uses their traces as contextual hints guiding the next execution steps.
Example of Planner-Plan-Optimizer-Critic interactions is in Appendix ~\ref{app:framework}.

\balance

\section{Ablation \texorpdfstring{\&}{and} Case Studies}
\subsection{Evaluation Setup}
\label{sec:5.1}
We evaluate our design on 286 real-world GUI tasks: 100 in-house, 116 from WebVoyager~\cite{he2024webvoyager} (643 total), and 70 from ScreenAgent~\cite{niu2024screenagent} (70 sessions). Tasks span local applications, web platforms, productivity tools, and cross-application workflows.

For task mining, we collect 30 active hours of interaction traces across 18 sessions, yielding 113 trace files. The mined corpus is intentionally distribution-shifted from the test suite: traces skew toward Local/App, while the evaluation set contains more Web/Crossover tasks. The traces cover 36 domains, whereas the 286-task suite spans 63 domains; only 22 overlap (34.92\%), leaving 41 unseen test domains (65.07\%). This setup stresses generalization of mining and retrieval rather than memorization. 
Domain distributions are detailed in Appendix~\ref{domaindist}.

All agents use the same atomic GUI action interface and identical timeout policies. We report task success (74.83\%), average completion ratio (91.14\%), Step Efficiency Ratio(successful steps / actual execution steps; higher is better). Differences are reported in percentage points (pp).

\subsection{End-to-End Execution of Ablated Models}

We ablate each component to analyze its impact on long-horizon planning stability and execution depth.

\begin{table}[t]
    \centering
    \caption{Component-wise ablation on planning. We report task success (\%) and averaged plan completion (\%).}
    \begin{tabular}{lcc}
        \toprule
        Method & Success (\%) $\uparrow$ & Completion (\%) $\uparrow$\\
        \midrule
        $B$ & 22.73 & 33.78\\
        $B + T_g$ & 46.43 & 57.41 \\
        $B + T_{SG} + Z$ & 54.64 & 77.56 \\
        $B + T_{SG} + S_{SG} + PM$ & 53.85 & 81.23 \\
        $B + T_{SG} + Z + PM$ & 62.51 & 85.00\\
        $B + T_{SG} + Z + S_{SG} + PM$ & \textbf{74.83} & \textbf{91.14}\\
        \bottomrule
    \end{tabular}
    \label{tab:ablation_success_only}
\end{table}

For Table~\ref{tab:ablation_success_only}, Completion denotes averaged plan completion 
(executed steps / synthesized plan steps), measuring execution progress beyond binary success.

We denote components as:
\textbf{$B$} (baseline planner–executor),
\textbf{$T_g$, $T_{SG}$} (greedy vs.\ $IG/SG$-gated trace retrieval),
\textbf{$Z$} (intent-level representation),
\textbf{$S_{SG}$} (subgroup skill hints),
\textbf{$PM$} (plan memory reuse)

Starting from \textbf{$B$} (22.73\% success), adding greedy trace retrieval (\textbf{$B$+$T_g$}) improves success by +23.7, pp showing that user traces substantially reduce cold-start errors but still induce drift in long horizons. Replacing greedy retrieval with $IG/SG$ gating + $Z$ brings a further +8.21, pp success gain and +20.15, pp completion gain, showing that organizing traces into representation-learned intent subgroups significantly stabilizes long-horizon execution.
Even without \textbf{$Z$}, combining \textbf{$S_{SG}$ + $PM$} achieves a high task completion rate (81.23\%), indicating that skill hints and plan memory alone substantially improve execution depth.

The full system (\textbf{$B$+$T_{SG}$+$Z$+$S_{SG}$+$PM$}) achieves the best success and completion overall, demonstrating that representation learning, intent-level gating, skill abstraction, and plan reuse act complementarily. 
Evaluation on step-wise planning consistency of the full system is in Appendix ~\ref{planning}.

\subsection{Case Study}

To illustrate the framework, we consider a task where the user asks the system to summarize a previously viewed lightweight-ML video and record the result in a personal workspace. This requires retrieval, reasoning, and coordination across multiple applications.

\begin{figure*}[t]
    \centering
    \includegraphics[width=.90\textwidth]{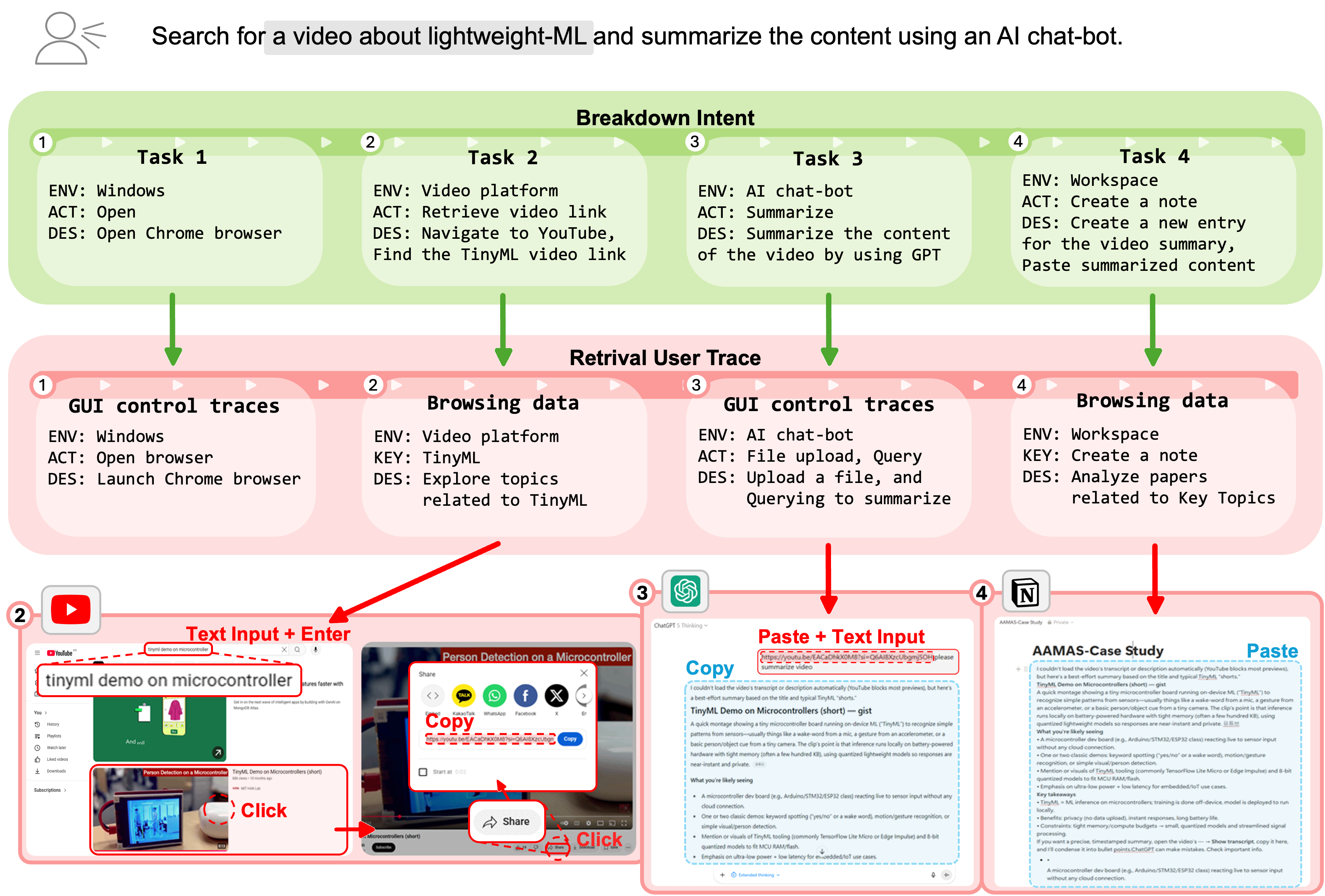}
    \caption{IntentCUA in action: the system recalls intent units from memory and decomposes a multi-application command into intent-level plan units, each executed through learned skills and recomposed into an end-to-end automation plan.}
    \label{fig:case_study}
\end{figure*}

Figure~\ref{fig:case_study} shows the execution process. The Planner retrieves relevant traces from plan memory, including prior interactions with the video platform, AI chatbot, and workspace application, and reconstructs the video source from the user’s history. 
It decomposes the request into structured intent-level plan units, each grounded into executable GUI actions by the Plan-Optimizer, while the Critic monitors progress and handles local inconsistencies. Through hierarchical reasoning and memory-guided skill retrieval, the system completes the multi-application task while maintaining intent coherence.

A failure case occurs when an unexpected pop-up appears during execution. Because underlying components become occluded, the grounding module fails to detect them, leading to incorrect retries. This highlights a limitation of script-based GUI grounding under transient interface changes.

\section{Experiments}
We compare \textbf{IntentCUA} against two representative desktop GUI agents chosen for methodological diversity:
UI-TARS-1.5~\cite{ui-tars-15-seed}, an RL-based visual planner–executor with self-evolving policies and screen grounding, and
UFO$^2$~\cite{zhang2025ufo2}, a trajectory-centric Windows automation agent that organizes demonstrations as executable sequences.
Together these baselines span reinforcement learning–driven automation versus demonstration-driven planning, and both operate at the level of atomic GUI actions, ensuring comparability with our interface.
We evaluate 286 tasks (the same evaluation suite described in Section~\ref{sec:5.1}) and report task success rate, Step Efficiency Ratio(SER), and Latency, further analyzing robustness by step-length bins, each step defined as a atomic action performed by the agent. SER is defined as the ratio of successful steps to total steps, ranging from 0 to 1.
Latency is measured as the execution time per task, reflecting not only the number of steps but also the overhead of perception and planning.

\subsection{Robust Long-Horizon Planning Efficiency}

\begin{table}[b]
\centering
\caption{End-to-end success rate comparison across datasets (\%). Columns show WebVoyager, ScreenAgent, our in-house suite, and overall average.}
\resizebox{\linewidth}{!}{
\begin{tabular}{lcccc}
    \toprule
    Method & WebVoyager & ScreenAgent & Ours & Total(\%) \\
    \midrule
    UI-TARS-1.5~\cite{ui-tars-15-seed} & 35.9 & 42.9 & 46.0 & 38.8 \\
    UFO$^2$~\cite{zhang2025ufo2}       & 69.0 & 41.4 & 38.0 & 51.2 \\
    \midrule
    \textbf{IntentCUA (ours)}          & \textbf{71.6} & \textbf{77.1} & \textbf{78.0} & \textbf{74.8} \\
    \bottomrule
\end{tabular}}
\label{tab:e2e_main}
\end{table}

\begin{figure}[t]
    \centering
    \includegraphics[width=0.95\linewidth]{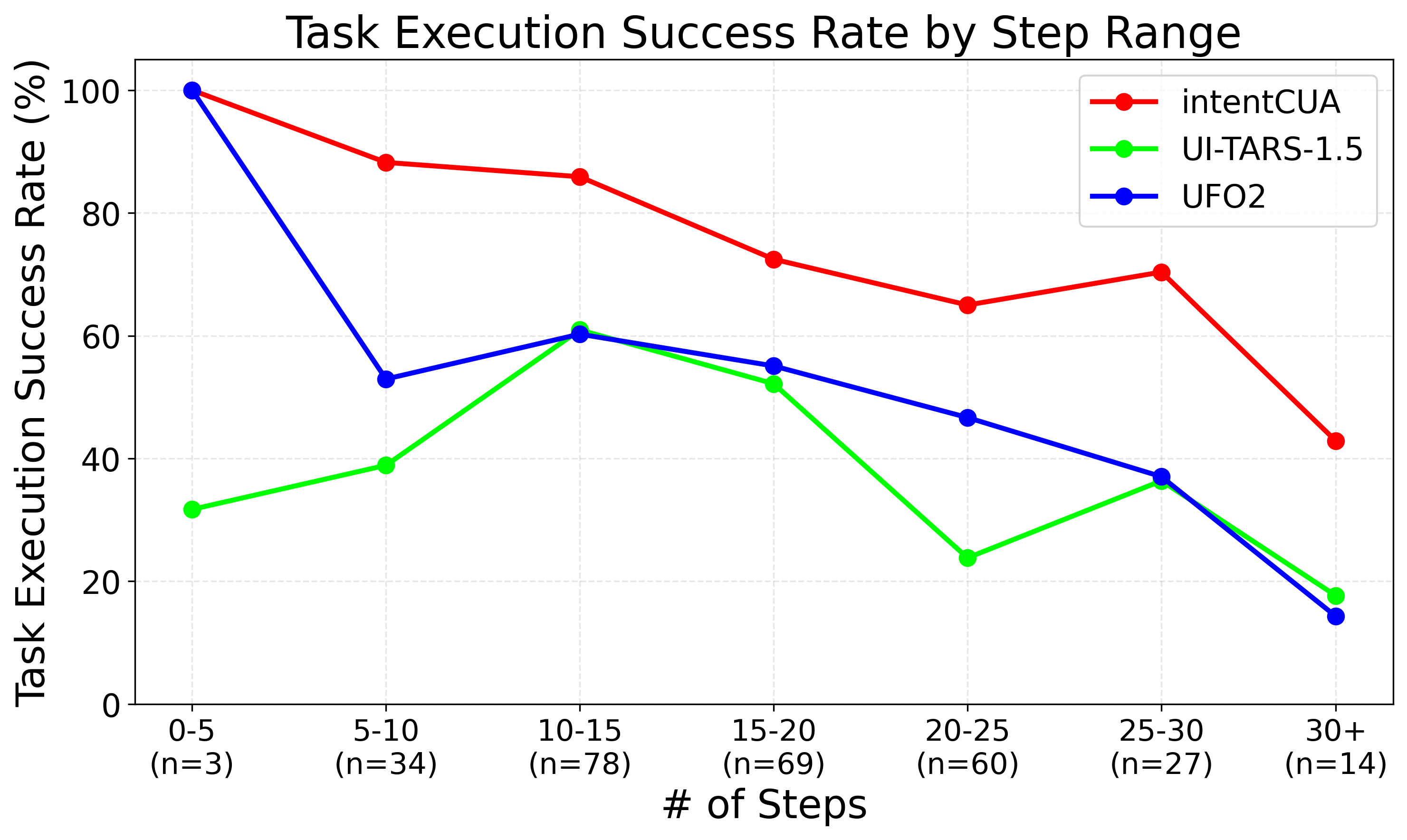}
    \caption{Success rate by step length (bin size = 5 steps). The x-axis shows step-length bins and the y-axis shows task success rate (\%).}
    \label{fig:success_bins}
\end{figure}

We evaluate how each agent sustains task completion as sequence length increases, focusing on the robustness of long-horizon planning.
Table~\ref{tab:e2e_main} and Figure~\ref{fig:success_bins} summarize overall and step-wise success trends across 286 evaluation tasks.
IntentCUA achieves the highest overall success rate of \textbf{74.8\%}, compared to 51.2\% for UFO$^2$ and 38.8\% for UI-TARS-1.5, yielding relative improvements of about +23.6 and +36 percentage points, respectively.

Notably, IntentCUA performs consistently well across all datasets, achieving 71.6\% on the web-based WebVoyager, 77.1\% on the cross-application ScreenAgent, and 78.0\% on our in-house local suite, demonstrating that its advantage is not confined to a specific benchmark.
While agents like UFO$^2$ specialize in narrow domains such as web navigation, IntentCUA generalizes effectively to heterogeneous desktop environments that include both online and offline contexts, confirming its versatility and domain robustness.

As shown in Figure~\ref{fig:success_bins}, IntentCUA maintains stable performance even as task length grows: 85.9\% at 10–15 steps, 72.5\% at 15–20, and 65.0\% at 20–25, while still retaining 42.9\% beyond 30 steps.
Both baselines, in contrast, decline sharply after 20 steps, dropping below 20\%.
This gradual degradation indicates that IntentCUA’s planning remains consistent and resistant to drift even in extended workflows spanning multiple windows and applications.

The stability across longer horizons can be attributed to its \emph{intent-aware retrieval} and \emph{plan memory reuse}, which enable the planner to recall previously successful subplans aligned with the current intent embedding rather than regenerating them from scratch.
Together, these results confirm that IntentCUA achieves robust and generalizable long-horizon planning efficiency, effectively preserving goal coherence and minimizing redundant re-planning under complex, real-world desktop environments.

\begin{figure}[b]
    \centering
    \includegraphics[width=0.95\linewidth]{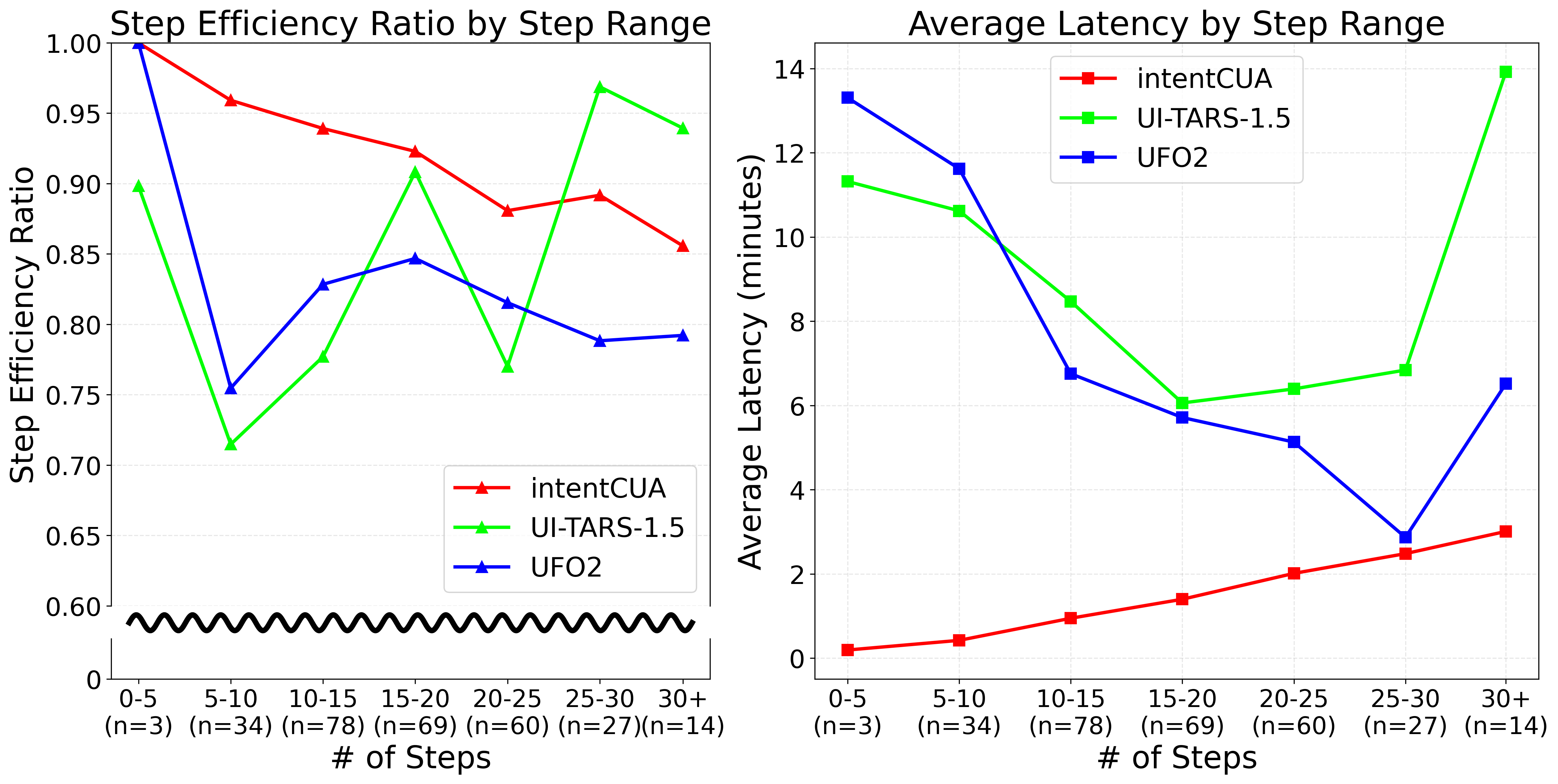}
    \caption{Performance by task length (bin size = 5 steps). Left: Step Efficiency Ratio (SER). Right: Average latency per task (minutes)}
    \label{fig:ser_lat_bins}
\end{figure}

\subsection{Stable \& Scalable Planning Efficiency and Latency}
We examine efficiency using two complementary metrics: the Step Efficiency Ratio (SER; Left) and the average latency per task (Right), as shown in Figure~\ref{fig:ser_lat_bins}.  
IntentCUA achieves the highest SER of \textbf{0.91}, exceeding UI-TARS (0.85) and UFO$^2$ (0.82).  
While SER in IntentCUA decreases moderately from 0.93 at 10–15 steps to 0.88 at 20–25, it remains consistently above 0.85 even for the longest tasks, indicating that most actions continue to contribute effectively to progress.  
In contrast, both baselines show sharper declines across similar ranges, suggesting increased redundancy or re-planning.

Latency patterns further highlight scalability.  
IntentCUA’s average execution time is 1.46 minutes, approximately 4.5× lower than the baselines (UFO$^2$: 6.63 min, UI-TARS: 9.82 min).  
Its latency increases smoothly with task length—for instance, from 0.95 min at 10–15 steps to 2.01 min at 20–25—showing near-linear growth.  
By comparison, UI-TARS exhibits irregular delays that expand sharply with step count, and UFO$^2$ shows unstable spikes on shorter tasks due to looped retries.

These results demonstrate that IntentCUA sustains high planning efficiency and low, predictable latency as task complexity increases.  
Its memory-guided retrieval and feedback design minimize redundant computation, yielding a scalable and robust planning policy suitable for real desktop automation.

\section{Conclusion}
We presented \textbf{IntentCUA}, a framework that transforms raw interaction traces into multi-view intent representations, abstracts them into reusable skills, and integrates these with plan memory to support stable long-horizon desktop automation. The system combines representation learning, hierarchical skill induction, and memory-guided planning to reduce re-planning and improve stability across complex workflows.

In experiments, IntentCUA achieved a 74.8\% task success rate with a step efficiency ratio of 0.91, outperforming both UI-TARS-1.5 (RL-based) and UFO$^2$ (trajectory-centric) by 4.5$\times$ times reduced latency. 
It also maintained over 40\% success on long-horizon tasks exceeding 30 steps. 
Ablation studies show that each component contributes to robustness and efficiency, with the full design providing the greatest improvements on longer tasks.
While IntentCUA shows consistent reasoning and cross-application generalization, several aspects remain open for refinement.
Retrieval efficiency may fluctuate as the plan memory grows, though this mainly affects latency rather than accuracy.
Graph-based retrieval and lightweight vision cues could further enhance robustness, allowing the system to adapt more smoothly to dynamic and visually changing interfaces.

\begin{acks}
This work was supported by the National Research Foundation of Korea(NRF) grant (No. RS-2022-NR066631, No. RS-2025-02216282) and Institute of Information \& communications Technology Planning \& Evaluation (IITP) grant (No.RS-2022-II220025) funded by the Korea government(MSIT) and Ministry of Trade, Industry and Energy of Korea (MOTIE RS 2023 00258591).
\end{acks}


\bibliographystyle{ACM-Reference-Format} 
\bibliography{sample}

\appendix
\section{Encoder Details}
\label{app:encoder}

This appendix summarizes the implementation details of the multi-view encoder described in Section~\ref{sec:3.1}.

\paragraph{Encoder Architecture.}
Each view (ENV, ACT/KEY, DES) is embedded using OpenAI \texttt{text-embedding-3-large}, producing a 3072-dimensional vector.
Each embedding is mapped to a shared latent space via a view-specific 2-layer MLP projection head:
\[
3072 \rightarrow 256 \rightarrow 256,
\]
with GeLU activation, dropout ($p=0.05$), and LayerNorm, yielding
\[
z_i^{(v)} \in \mathbb{R}^{256}.
\]

Cross-view consistency is enforced using six symmetric dual predictors, one for each ordered view pair.
Each predictor is a lightweight MLP ($256 \rightarrow 128 \rightarrow 256$) used only during training.
Additionally, a linear decoder ($256 \rightarrow 3072$) is applied per view to reconstruct the original embedding for reconstruction regularization.

\paragraph{Shared Representation and Fusion Weights.}
The final shared intent representation is computed as a weighted fusion:
\[
z_i = 0.4\,z_i^{(E)} + 0.3\,z_i^{(A)} + 0.3\,z_i^{(D)}.
\]
The environment view is assigned the largest weight because execution environment provides the most stable contextual signal in desktop automation and serves as the primary driver for upper-level intent group (IG) formation.
This environment-centric fusion improves the stability of hierarchical clustering while still preserving action- and description-level variability for finer subgroups (SG).

For model training, we used a learning rate of $1\times 10^{-3}$, with $\lambda_{\mathrm{pred}}=0.1$, $\lambda_{\mathrm{rec}}=0.05$, and a contrastive temperature $\tau=0.1$.

\paragraph{Tensor Shapes.}
For a minibatch of size $N$, the encoder operates on:
\[
x^{(v)} \in \mathbb{R}^{N \times 3072}, \quad
z^{(v)} \in \mathbb{R}^{N \times 256}, \quad
z \in \mathbb{R}^{N \times 256}.
\]

\section{Framework Details (Planner-Plan-Optimizer-Critic)}
\label{app:framework}

\begin{figure*}[t]
    \centering
    \includegraphics[width=.90\textwidth]{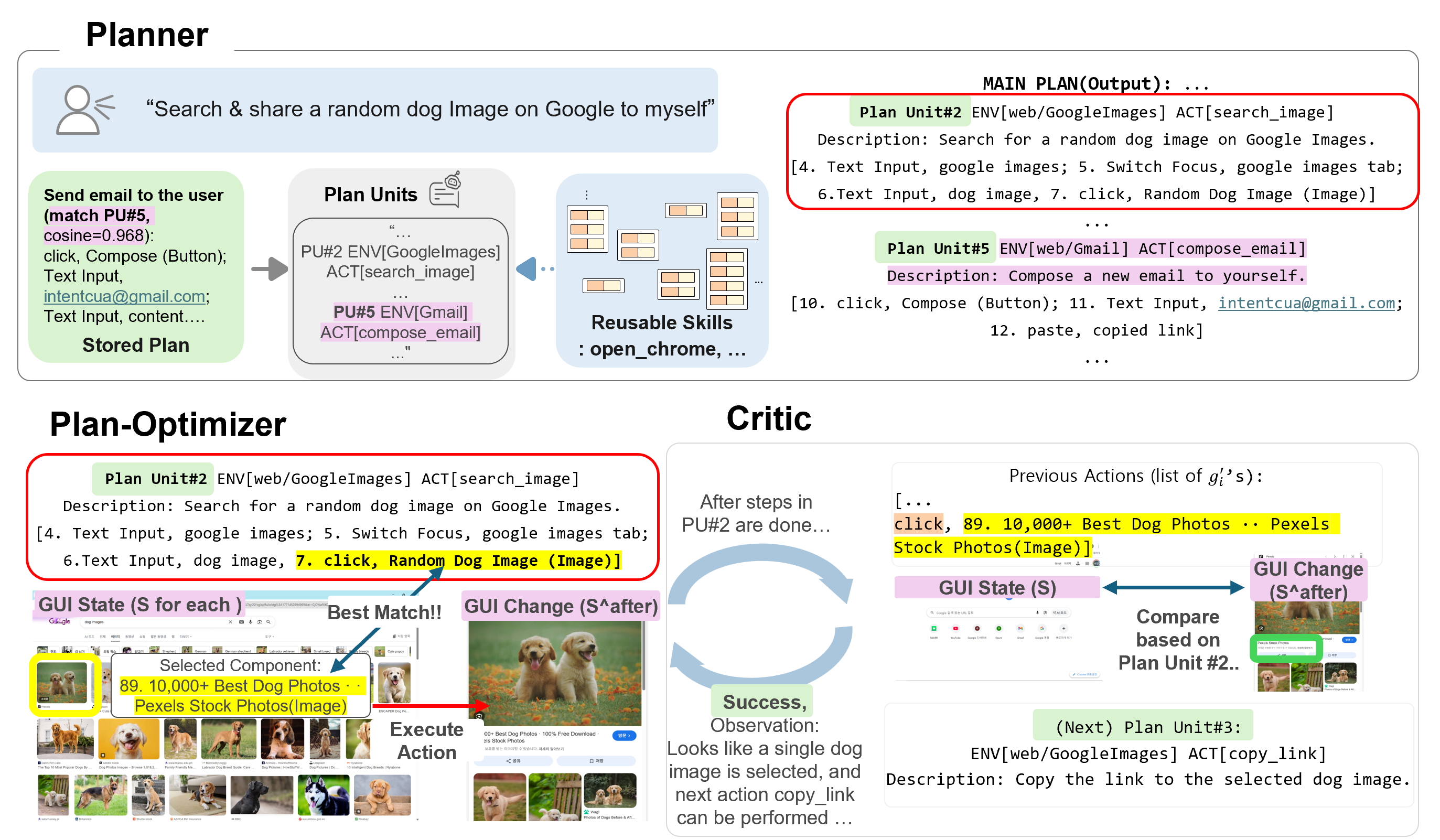}
    \caption{Planner–Plan-Optimizer–Critic interaction. The Planner decomposes a command into structured Plan Units and retrieves or synthesizes a global plan. The Plan-Optimizer grounds each unit into executable GUI actions conditioned on the current state, while the Critic validates the post-state $S^{\text{after}}$ and triggers local re-optimization if needed.}
    \label{fig:interaction}
\end{figure*}

\paragraph{Planner}
At inference time, the Planner maps a natural-language command $c$ into the same structured format used for log labeling by prompting the LLM to produce task units of the form Task Unit: ENV[...] ACT[...], Task k: ENV[...] ACT[...] with short descriptions (e.g., “search dog at a browser” → ENV[local/Windows, web/searching browser], ACT[open browser, search]). 

From these views, it builds an intent prototype and retrieves candidate plans from memory. A cached plan is reused only if its action coverage with respect to the current breakdown is high: in practice, we require that the plan already contains most of the required ACTs (allowing at most 2 missing ACTs per command). 
For each such missing ACT, we retrieve the corresponding intent subgroup, select the skill template most frequently observed in the logs, instantiate its placeholders from $c$ (e.g., {query} = "dog"), and splice the resulting steps into the cached plan. If no cached plan satisfies this condition, the Planner falls back to RAG using representative logs as examples. The final output is a global plan G = ${g_i}$, where each step $g_i$ = (action, object) uses one of a fixed set of 17 low-level GUI actions (e.g., text input, click, doubleclick, press, switch focus, save, copy…).

\paragraph{Execution(Plan-Optimizer → Critic interaction)}
Execution consists of a Plan-Optimizer that grounds each $g$ into concrete GUI actions, and a Critic that validates the post-state and triggers local recovery. Each step $g$ from the global plan is expanded into an actionable sequence $g'$ using a fixed library of default action templates (e.g., open : {doubleclick icon or click taskbar → type target →press enter}), ensuring grounding into atomic GUI actions.

The Plan-Optimizer conditions on (1) task-unit context, (2) the parsed screen state $s$, and (3) a retrieved plan hint by matching $z^{(pu)}$ to the nearest subgroup centroid. Hints contain historical {ENV/ACT/DES} tuples and action-object traces, biasing execution toward stable patterns rather than free-form generation.

After each plan unit, the Critic inspects the post-state $s^{\text{after}}$ via a structured prompt that checks window focus, component availability, and compatibility with the next expected step and returns retryable={success, retryable, blocked} which triggers a localized re-optimization from $s^{\text{after}}$, avoiding global re-planning, while ‘blocked’ indicates that neither template-based execution nor exemplar-guided adjustment provides a safe continuation. 
A step-by-step example of this interaction is illustrated in Figure~\ref{fig:interaction}.

\section{Ablation on the Representation Loss}
\label{ablation}

\begin{table}[t]
\centering
\caption{Representation loss ablation on intent embedding quality.
We report size-weighted density separation (inter/intra) over all HDBSCAN subgroups.}
\begin{tabular}{lc}
\toprule
Loss Variant & Separation $\uparrow$\\
\midrule
Baseline Embedding &5.60 \\
InfoNCE only &5.64 \\
InfoNCE + Prediction &6.92 \\
InfoNCE + Reconstruction &23.17 \\
\midrule
\textbf{Full (Con + Pred + Rec)} &\textbf{7.74} \\
\bottomrule
\end{tabular}
\label{tab:repr_loss_ablation}
\end{table}

\begin{table}[t]
\centering
\caption{ENV/ACT purity under different representation loss variants.
We report mean purity with standard deviation in parentheses.}
\begin{tabular}{lcc}
\toprule
Loss Variant & ENV purity & ACT purity \\
\midrule
Baseline Embedding & 0.83(0.19) & 0.48(0.25)\\
InfoNCE only & 0.82 (0.22) & 0.37 (0.22) \\
InfoNCE + Prediction & 0.83 (0.21) & 0.42 (0.26) \\
InfoNCE + Reconstruction & 0.86 (0.19) & 0.42 (0.26) \\
\midrule
\textbf{Full (Con + Pred + Rec)} & \textbf{0.84 (0.20)} & \textbf{0.42 (0.23)} \\
\bottomrule
\end{tabular}
\label{tab:purity_ablation}
\end{table}

We analyze density separation and cluster purity with respect to ENV and ACT tags for further ablation.

To address concerns about the representation objective, we compare four variants: 
InfoNCE-only, InfoNCE + cross-view prediction, InfoNCE + reconstruction, and the full loss. 

\begin{equation}
\mathrm{Sep}_{\mathrm{w}}
=
\frac{\sum\limits_{IG}\sum\limits_{SG\in IG}
|SG|\;
\frac{\mathrm{Inter}(SG)}{\mathrm{Intra}(SG)}
}
{\sum\limits_{IG}\sum\limits_{SG\in IG} |SG|}
\label{eq:sep}
\end{equation}

\begin{equation}
\begin{aligned}
\mathrm{Intra}(SG)
&=
\frac{1}{|SG|}
\sum_{x\in SG}
d(x,\mathbf{c}_{SG}) \\
\mathrm{Inter}(SG)
&=
\min_{SG' \neq SG}
d(\mathbf{c}_{SG}, \mathbf{c}_{SG'})
\end{aligned}
\label{eq:inter-intra}
\end{equation}

\begin{equation}
\begin{aligned}
\mu^{X}
&=
\frac{1}{|\mathcal{SG}|}
\sum_{IG}\sum_{SG\in IG}
\mathrm{Purity}^{X}(SG) \\
\sigma^{X}
&=
\sqrt{
\frac{1}{|\mathcal{SG}|}
\sum_{IG}\sum_{SG\in IG}
\left(
\mathrm{Purity}^{X}(SG)-\mu^{X}
\right)^2
}
\end{aligned}
\label{eq:sg-purity-mean}
\end{equation}

\begin{equation}
\mathrm{Purity}^{X}(SG)
=
\frac{
\max_{c}
\left|
\{\, t\in SG : \mathrm{tag}^{X}(t)=c \}
\right|
}
{|SG|}
\label{eq:purity}
\end{equation}

We report 
(1) size-weighted density separation (inter/intra: Eq ~\ref{eq:inter-intra}) in Eq ~\ref{eq:sep} over all HDBSCAN subgroups in Table ~\ref{tab:repr_loss_ablation} and 
(2) semantic purity(Eq ~\ref{eq:sg-purity-mean}) measured as majority ENV/ACT ratios(Eq ~\ref{eq:purity}) within each subgroup in Table ~\ref{tab:purity_ablation}.

The full loss improves the separation ratio from 5.64 (InfoNCE-only) to 7.74, while 
InfoNCE + reconstruction produces an inflated separation score of 23.17 due to extreme micro-clusters (size=2), 
indicating over-fragmentation rather than robust intent abstraction. 
In terms of semantic consistency, ENV purity increases from 0.82 (InfoNCE-only) to 0.84 (Full), 
and ACT purity improves from 0.37 to 0.42. 
Reconstruction yields the highest ENV purity (0.86) but with higher fragmentation, 
while Prediction consistently improves ACT purity (0.42 vs. 0.37 in InfoNCE-only). 
The full objective maintains balanced ENV/ACT purity (0.84 / 0.42) with reduced variance 
(ENV std 0.20, ACT std 0.23), suggesting more stable and semantically coherent intent embeddings.

\section{Domain level distributions of dataset/testcases}
\label{domaindist}

\begin{figure}[b]
    \centering
    \includegraphics[width=\linewidth]{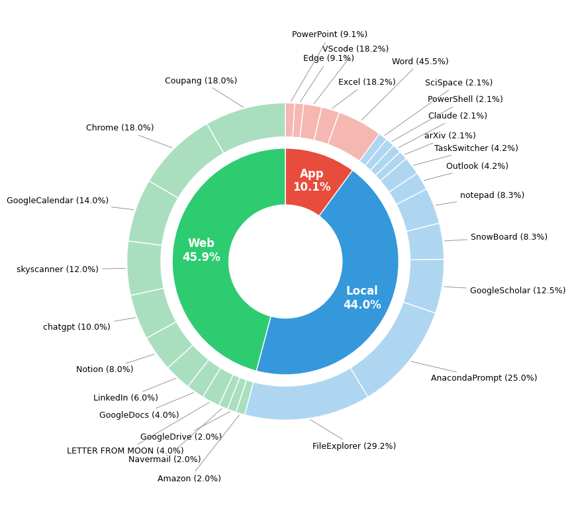}
    \caption{Domain distribution of collected trace data. Each slice indicates a domain category and its proportion within the trace corpus (\%).}
    \label{fig:data_dist}
\end{figure}

\begin{figure}[b]
    \centering
    \includegraphics[width=\linewidth]{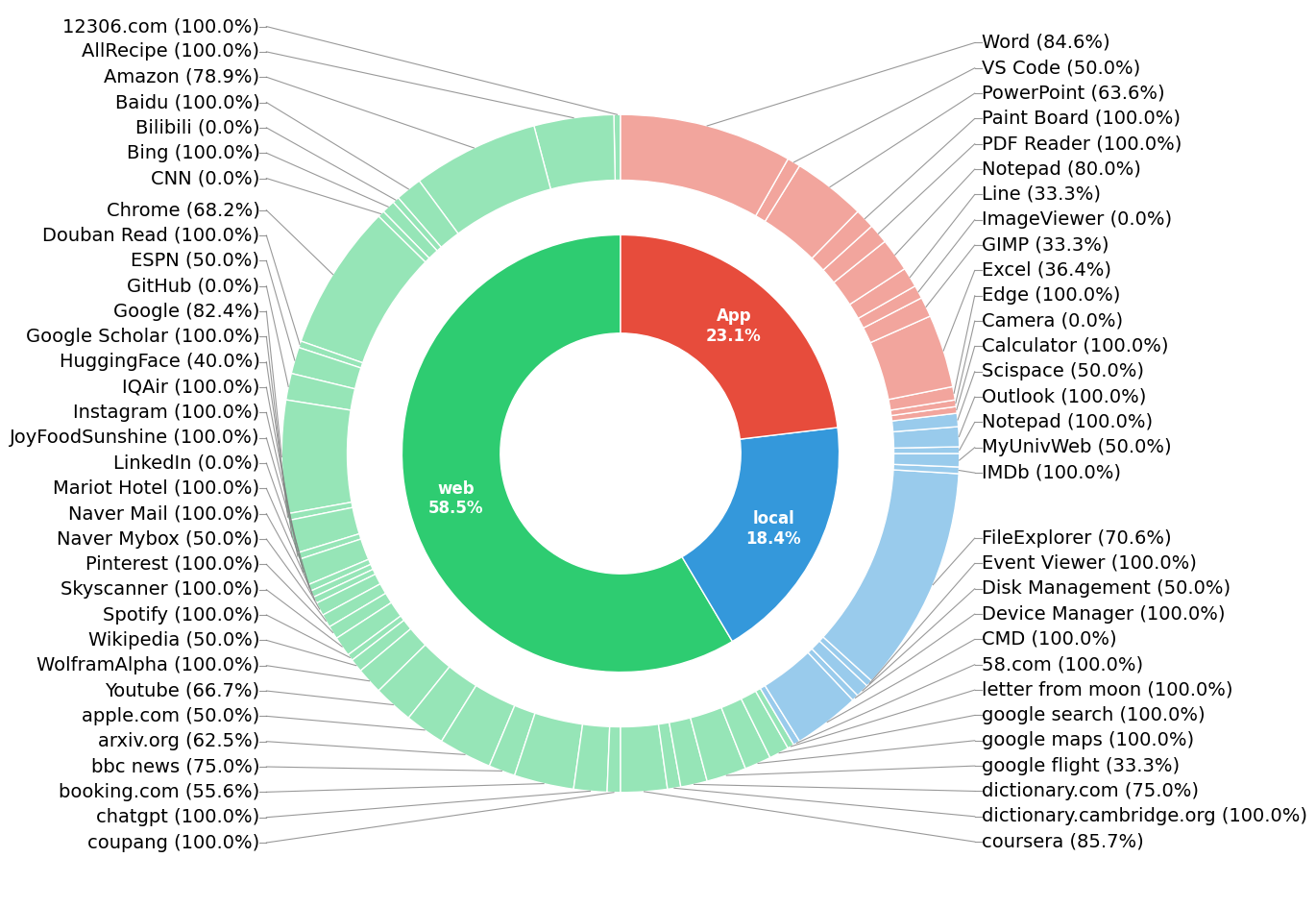}
    \caption{Domain distribution of the 286 evaluation testcases. Each slice shows a domain category and the success rate achieved within that domain (\%).}
    \label{fig:testcase_dist}
\end{figure}

Figure~\ref{fig:data_dist} shows the domain distribution of the collected trace corpus. The trace data are skewed toward Local/App environments, with several long-tail domains having only a few interaction sessions. In contrast, the evaluation suite (Figure~\ref{fig:testcase_dist}) contains a broader Web/Crossover share and substantially more domains overall. This asymmetry reflects the intentional distribution shift described in the main text.

Importantly, domain-level success rates in Figure~\ref{fig:testcase_dist} indicate that performance does not strictly correlate with trace frequency. Several domains with very limited or no traces still achieve non-trivial success rates, suggesting that the planner generalizes beyond memorized trajectories. While trace sparsity and bias remain limitations, these statistics provide additional transparency regarding domain coverage and generalization behavior.

\section{Planning Consistency}
\label{planning}

To further assess stability under system complexity, we introduce \textit{step consistency} as a quantitative measure of plan repeatability. 
For each of the 286 testcases, we execute planning five times and compare the resulting plans. 
For each plan unit $pu$, we examine whether the generated step sequence is consistently reproduced across all five runs. 
A $pu$ is counted as consistent if the pairwise cosine similarity between corresponding step embeddings exceeds a threshold of 0.93 in all comparisons. 
Step consistency is defined as the proportion of such consistent $pu$ instances within each benchmark split.

Table~\ref{tab:planning_consistency} reports the mean and standard deviation across domains. 
The results indicate stable planning behavior across datasets, including unseen domains. 
Despite distribution shift and sparse traces, the Planner–Plan-Optimizer loop maintains high repeatability, suggesting that structural constraints (e.g., ENV derived from window structure and ACT from UI semantics) effectively reduce LLM drift and labeling variance.

\begin{table}[t]
\centering
\caption{Step consistency (\%) across five repeated planning runs per testcase (cosine threshold = 0.93).}
\label{tab:planning_consistency}
\begin{tabular}{lcccc}
\toprule
 & WebVoyager & ScreenAgent & Ours & Total \\
\midrule
Mean (\%) & 70.3 & 81.1 & 86.2 & 78.5 \\
STD (\%)  & 8.2  & 5.3  & 7.9   & 7.8  \\
\bottomrule
\end{tabular}
\end{table}


\end{document}